\documentclass[10pt,twocolumn,letterpaper,twoside]{article} 

\usepackage{comment}
\usepackage{amsmath}
\usepackage{amsfonts}
\usepackage[headings,in]{fullpage}
\usepackage[dvips]{graphicx}
\usepackage{psfrag}
\usepackage[nice]{nicefrac}
\usepackage{newcent}		
\usepackage[round]{natbib}
\usepackage{url}
\DeclareMathOperator{\var}{var}
\DeclareMathOperator{\stderr}{stderr}
\newcommand{\bigfrac}[2]{\frac{\displaystyle #1}{\displaystyle #2}}
\newcommand{\ifrac}[2]{({#1}/{#2})}

\newcommand{\eq}[1]{Eq.~\ref{eq:#1}}
\newcommand{\fig}[1]{Fig.~\ref{fig:#1}}
\DeclareMathOperator{\noise}{\mathcal{N}}

\newlength{\gwidth}
\graphicspath{
  {figures/}
  {/home/barak/src/papers/query-max/figures/}
  {figs/}
  {figs/home/barak/src/papers/query-max/figures/}
  {./}				
}
\DeclareGraphicsExtensions{.eps,.eps.gz,.jpg,.gif,.png}
\title{Bounds on Query Convergence}

\author{\textbf{Barak A. Pearlmutter}\thanks{Hamilton Institute, NUI
Maynooth, Co.\ Kildare, Ireland.}}

\date{\today\\\small (CVS: \rcsInfoFile\ \rcsInfoRevision)}
\date{}

\begin{document}
\maketitle
\thispagestyle{empty}

\begin{abstract}
The problem of finding an optimum using noisy evaluations of a smooth
cost function arises in many contexts, including economics, business,
medicine, experiment design, and foraging theory.  We derive an
asymptotic bound
\begin{math}
 E[ (x_t-x^*)^2 ] \geq O(t^{-1/2})
\end{math}
on the rate of convergence of a sequence $(x_0, x_1, \ldots)$
generated by an unbiased feedback process observing noisy evaluations
of an unknown quadratic function maximised at $x^*$.  The bound is
tight, as the proof leads to a simple algorithm which meets it.  We
further establish a bound on the total regret,
\begin{math}
 E\bigl[ \sum_{\tau=1}^{t} (x_{\tau} - x^*)^2 \bigr] \geq O(t^{1/2}) .
\end{math}
These bounds may impose practical limitations on an agent's
performance, as $O(\epsilon^{-4})$ queries are made before the queries
converge to $x^*$ with $\epsilon$ accuracy.
\end{abstract}

\section{Introduction}

Finding an input $x$ to a system so as to optimise some property
$f(x)$ of the system's output, using only noisy measurements, is a
ubiquitous problem.  For instance, in medicine $x$ might be a drug
dosage and $f(x)$ the probability of a successful outcome; in business
$x$ might be the price set by a manufacturer and $f(x)$ the consequent
profit; in game theory $x$ might be a strategy and $f(x)$ its return;
and in evolutionary theory $x$ might be the brightness of a bird's
plumage and $f(x)$ the consequent reproductive success.

When the measurements of $f(x)$ are noise-free this is a classical
optimisation problem, as studied by Gauss.  Optimisation theory
remains to this day a productive branch of applied mathematics.  In
general, the assumption is made that the function to be optimised
takes on a simplified form in the neighbourhood of its optimum---most
often, quadratic.  The criterion by which we evaluate such algorithms
is typically the convergence rate of its estimate of the location of
the optimum, although the complexity of the algorithm itself can also
be a consideration.

Here we consider a situation in which the measurements of the function
are assumed to be noisy.  A similar situation in which noisy
measurements of the gradient are available is studied in stochastic
gradient optimisation \citep{ROBBINS-MONRO51a, LJUNG77,
WIDROW-MCCOOL-LARIMORE-JOHNSON79}.  Here however we assume that
gradient information is not available.  We further assume that we are
interested not in our \emph{estimate} of the optimum converging as
rapidly as possible, but rather in the \emph{queries themselves}
converging to the optimum as rapidly as possible.  As a practical
matter, the convergence of the queries themselves is important when
the function $f(x)$ is a measure of consequence, and making a
measurement at $x$ has an actual expected cost of $f(x)$, as in
measuring the survival rate of a medical treatment or the return of an
economic decision.

Gradient information would make this problem much easier.  For
illustration, consider two closely related optimisation problems.  In
each, an inaccurate rifle with unknown bias can be swivelled
horizontally, and we wish to swivel it so as to maximise the
probability of hitting a small target.  Due to the inaccuracy of the
riffle and the small target size, we are unlikely to hit the target
even when the rifle is aimed optimally.  In one situation, we know
after each shot whether the bullet went to the left or the right of
the target.  In the other situation, we know only whether the bullet
hit the target.  Knowing whether the bullet went to the right or the
left of the target corresponds to having an estimate of the gradient,
and allows rapid convergence to the correct position by simply making
successively smaller adjustments after each shot away from the side to
which the bullet missed.  But without this gradient information, it is
difficult to know in which direction to adjust the aim in response to
a miss.  In fact, a single miss in isolation does not seem of any help
in improving the aim.  It is our goal here to precisely characterise
the difficulty of such situations.

\section{Proof Sketch}

We construct an inequality which establishes a lower bound on the rate
of convergence of the queries $x_t$ to the optimum $x^*$.  The
inequality follows from the observation that if the queries $x_t$ are
more spread out, the estimate of the optimum $x^*$ will have less
uncertainty.  This relationship, in which faster convergence of the
queries leads to slower convergence of the estimate of $x^*$, is
quantified using the statistical notion of the leverage of the data,
which limits the accuracy of an estimate of a slope.  This gives a
lower bound on the speed with which the queries $x_t$ can converge to
$x^*$.  Violation of the bound would imply a contradiction: that the
queries converge to the optimum faster than does the best estimate of
the optimum.

\section{Detailed Derivation}

We consider an unbiased feedback system which uses noisy measurements
to find the $x$ which maximises $f(x)$, where $f(x)$ is locally
quadratic about its maximum $x^*$.  To simplify the derivation we will
assume that $f(x)$ is not merely locally but globally quadratic
\begin{equation}
 f(x) = - a x^2 + b x + c = -a (x - x^*)^2 + f(x^*)
\end{equation}
that the quadratic coefficient $a>0$ is known leaving unknown only the
linear and constant terms $b$ and $c$, and that each noisy
measurements of $f(x)$ is corrupted by zero-mean i.i.d.\ additive
noise of variance $\sigma^2$.

Let $x_0, x_1, \ldots$ be the sequence of points evaluated.  We
establish the following bound:

\newtheorem{theorem}{Theorem}
\newtheorem{corollary}{Corollary}

\begin{theorem} \label{theorem:main}
For sufficiently large $t$ and an unbiased feedback process that
calculates $x_t$ using information available prior to $t$,
\begin{equation} \label{eq:main_thm} \displaystyle
  E[ (x_t - x^*)^2 ] \geq \frac{\sigma}{\sqrt{8} \, a} \, t^{-1/2}
\end{equation}
\end{theorem}

\textbf{Proof:}
Since $a$ is known we can add $a x_t^2$ to the measurements and fit
$b$ and $c$ to the resulting noisy line.  The variance of $\hat{b}_t$,
the best unbiased estimate of $b$ given measurements made prior to
time $t$, is limited by the Cram\'er-Rao bound which depends on the
level of measurement noise and the leverage about the sample mean
$\overline{x}_t = (x_0 + x_1 + \cdots + x_{t-1})/t$,
\begin{equation}
  \var \hat{b}_t
	= \bigfrac{ \sigma^2 }
		  { \sum_{\tau<t} (x_{\tau} - \overline{x}_t)^2 } .
\end{equation}

This leverage is bounded by the leverage about any point; here we
choose $x^*$, the desired point of convergence,
\begin{equation}
  \sum_{\tau<t} (x_{\tau} - \overline{x}_t)^2
	\leq \sum_{\tau<t} (x_{\tau} - x^*)^2
\end{equation}
so
\begin{equation}
  \var \hat{b}_t
	\geq \bigfrac{\sigma^2}{ \sum_{\tau<t} (x_{\tau} - x^*)^2 }
\end{equation}
Because $x^* = b/2a$ the variance of an estimate of $x^*$ is related to
the variance of an estimate of $b$,
\begin{equation}
  \var \hat{x}^*_t = \frac{1}{4a^2} \var \hat{b}_t
\end{equation}
where $\hat{x}^*_t$ is the best unbiased estimate of $x^*$ given
measurements made prior to $t$.  By definition $\hat{x}_t^*$ cannot be
a worse estimate of $x^*$ than is $x_t$, and we have already seen a
bound on the quality of the estimate $\hat{x}_t^*$, so
\begin{equation} \label{eq:two_sided}
  E[ (x_t - x^*)^2 ]
	\geq \var \hat{x}^*_t
	\geq \bigfrac{\sigma^2}{4a^2 \sum_{\tau<t} (x_{\tau} - x^*)^2 }
\end{equation}
where the expectation $E[\cdot]$ is taken over realisations of the
measurement noise.

We now assume\footnotemark\ that $x_t$ convergences polynomially,
$E[(x_t - x^*)^2] = (k t^r)^2$, and substitute this above to find $r$
and $k$.  The leverage about $x^*$ can be evaluated,
\footnotetext{If the fastest possible convergence bound were not of
this form then we would obtain a valid bound, but not a tight one.
However, we constructively show that the bound obtained is tight.}
\begin{equation} \label{eq:form}
  E\Bigl[ \sum_{\tau<t} (x_{\tau} - x^*)^2 \Bigr]
	= k^2 \sum_{\tau<t} \tau^{2r}
	= \frac{k^2}{1+2r} t^{1+2r}
\end{equation}
\eq{form} can be substituted into the two-sided bound on
$\var\hat{x}^*_t$ in \eq{two_sided}, yielding
\begin{gather}
     k^2 t^{2r}
     = E[ (x_t - x^*)^2 ]
     \geq \var \hat{x}^*_t
     \geq \frac{\sigma^2 (1+2r)}{4 k^2 a^2} t^{-(1+2r)}
\nonumber\\
\intertext{or}
  k^4 \geq \frac{\sigma^2 (1+2r)}{4a^2} t^{-(1+4r)}
\end{gather}
This can only be satisfied if the right hand side is bounded, which
implies that $r \geq -1/4$, and hence
\begin{equation}
  E[(x_t - x^*)^2] \geq O(t^{-1/2})
\end{equation}
The most aggressive convergence is for $r=-1/4$, at which point
equality is achieved when $k^2 = \sigma/(\sqrt{8} \, a)$.
Substituting yields \eq{main_thm}.

\begin{corollary}[Bound on Instantaneous Regret]
The expected instantaneous regret (loss incurred at time $t$ due to
ignorance) of an unbiased online optimiser is bounded below in
expectation by
\begin{equation}
 E[f(x^*) - f(x_t)] \geq \frac{\sigma}{\sqrt{8}} t^{-1/2}
\end{equation}
\end{corollary}

\textbf{Proof:} Note that $f(x^*) - f(x) = a (x - x^*)^2$ and
substitute into Theorem \ref{theorem:main}.

\begin{corollary}[Bound on Total Regret]
The total regret prior to time $t$, defined by
\begin{math}
 R_t = \sum_{\tau<t} f(x^*) - f(x_{\tau}) ,
\end{math}
incurred by an unbiased feedback process is bounded below in
expectation by
\begin{equation}
 E[R_t] \geq  \frac{\sigma}{\sqrt{2}} t^{1/2}
\end{equation}
\end{corollary}

\textbf{Proof:} Summation of the bound on instantaneous regret.

\textbf{Note:} The expected regret bound is independent of the
constant of curvature $a$, whose effect cancels itself out in the
analysis.  This is necessarily the case, because we could define
$\tilde{f}(x) = f(100 \, x)$ and an attempt to optimise $\tilde{f}(x)$
should yield the same regret as an attempt to optimise $f(x)$, despite
their differing curvatures.

\begin{theorem}[Optimal Algorithm] \label{thm:alg}
The stochastic algorithm
\begin{equation}
 x_t = \hat{x}^*_t + \noise\bigl((\stderr \hat{x}^*_t)^p\bigr)
\end{equation}
is unbiased and with $p=2$ achieves $E[(x_t - x^*)^2] \sim
\ifrac{\sqrt{2} \, \sigma}{a} \, t^{-1/2}$ and $E[R_t] \sim \sigma
\sqrt{8 t\,}$, where $\noise(\varsigma^2)$ is zero-mean
$\varsigma^2$-variance i.i.d.\ noise and $\stderr \hat{x}^*_t$ is the
standard error of the unbiased estimator $\hat{x}^*_t$.
\end{theorem}

\textbf{Proof:} The algorithm involves only unbiased estimates and is
therefore unbiased.

The inequalities above become equalities when
\begin{equation}
 x_t = \hat{x}^*_t + \noise\bigl(\sqrt{2} \, \sigma a \, t^{-1/2} \bigr)
\end{equation}
which has the same injected variance (up to absorbed constant factors)
as in the proposed algorithm.

\textbf{Note:} The existence of this algorithm implies that the
earlier bounds are tight.  Interestingly, the algorithm does not
require knowledge of $a$ or $\sigma$, which are used only in the
analysis.  Due to the statistics of the situation, $\stderr
\hat{x}^*_t$ scales appropriately with $a$ and $\sigma$.

\begin{figure*}[t!]
\psfrag{time}[c][c]{$t$}
\psfrag{R(t)}[c][c]{$R_t$}
\psfrag{ 0}[r][r]{0}
\psfrag{ 10000}[r][c]{$10^4$}
\psfrag{ 200}[r][r]{200}
\psfrag{ 500}[r][r]{500}
\psfrag{ 1000}[r][r]{1000}
\psfrag{no noise}[c][c]{Greedy: $x_t = \hat{x}^*_t$}
\psfrag{p=0.8}[c][c]{$x_t = \hat{x}^*_t + \noise((\stderr\hat{x}^*_t)^{0.8})$}
\psfrag{p=2}[c][c]{$x_t = \hat{x}^*_t + \noise((\stderr\hat{x}^*_t)^{2})$}
\psfrag{p=3.6}[c][c]{$x_t = \hat{x}^*_t + \noise((\stderr\hat{x}^*_t)^{3.6})$}
\includegraphics[width=\gwidth]{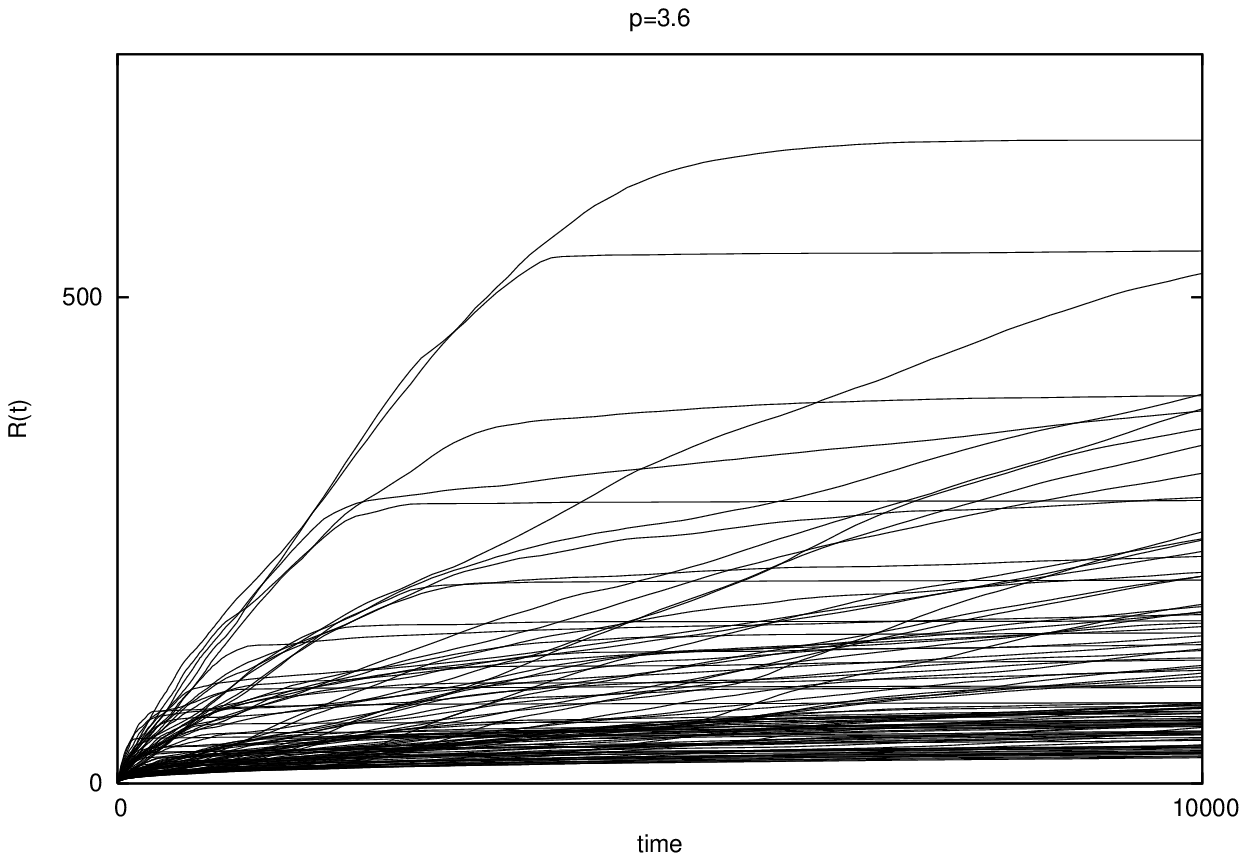}\hfill%
\includegraphics[width=\gwidth]{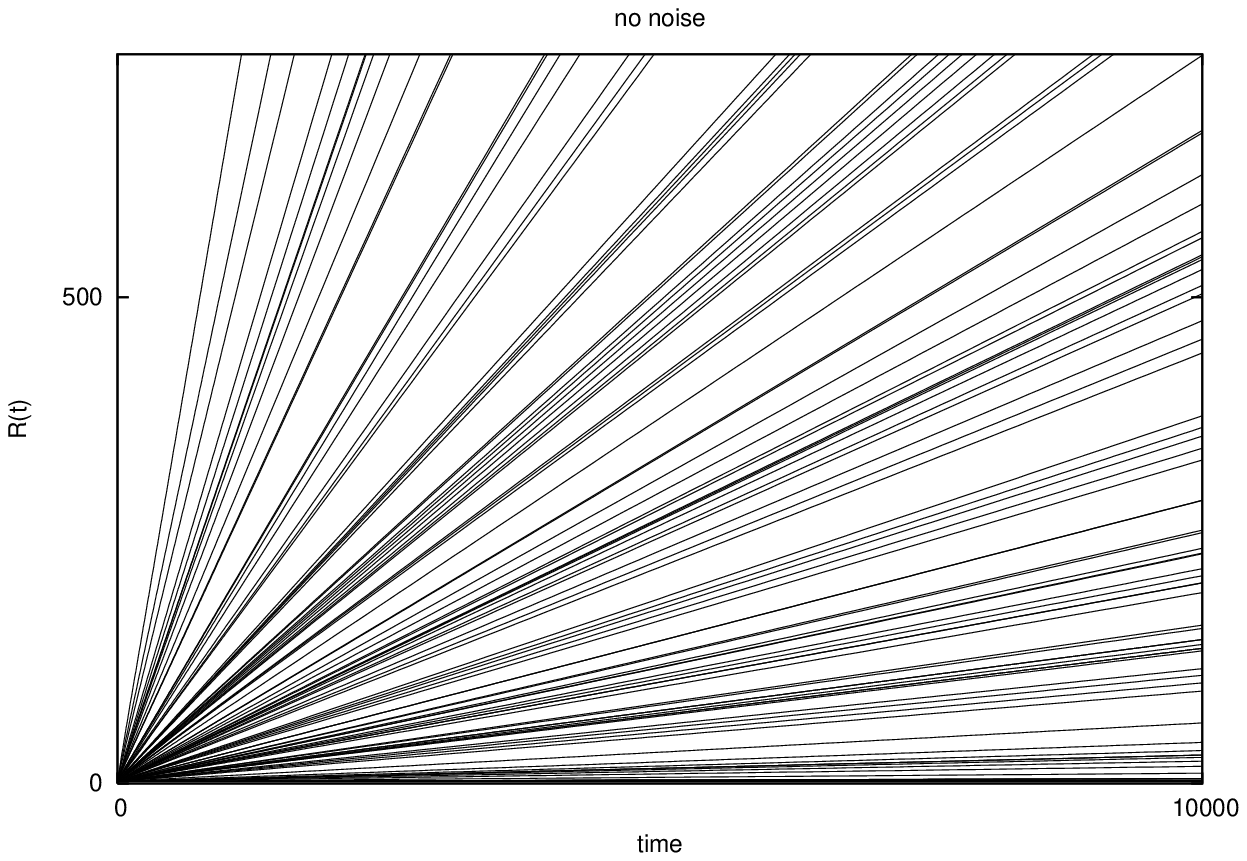}\\[2ex]
\includegraphics[width=\gwidth]{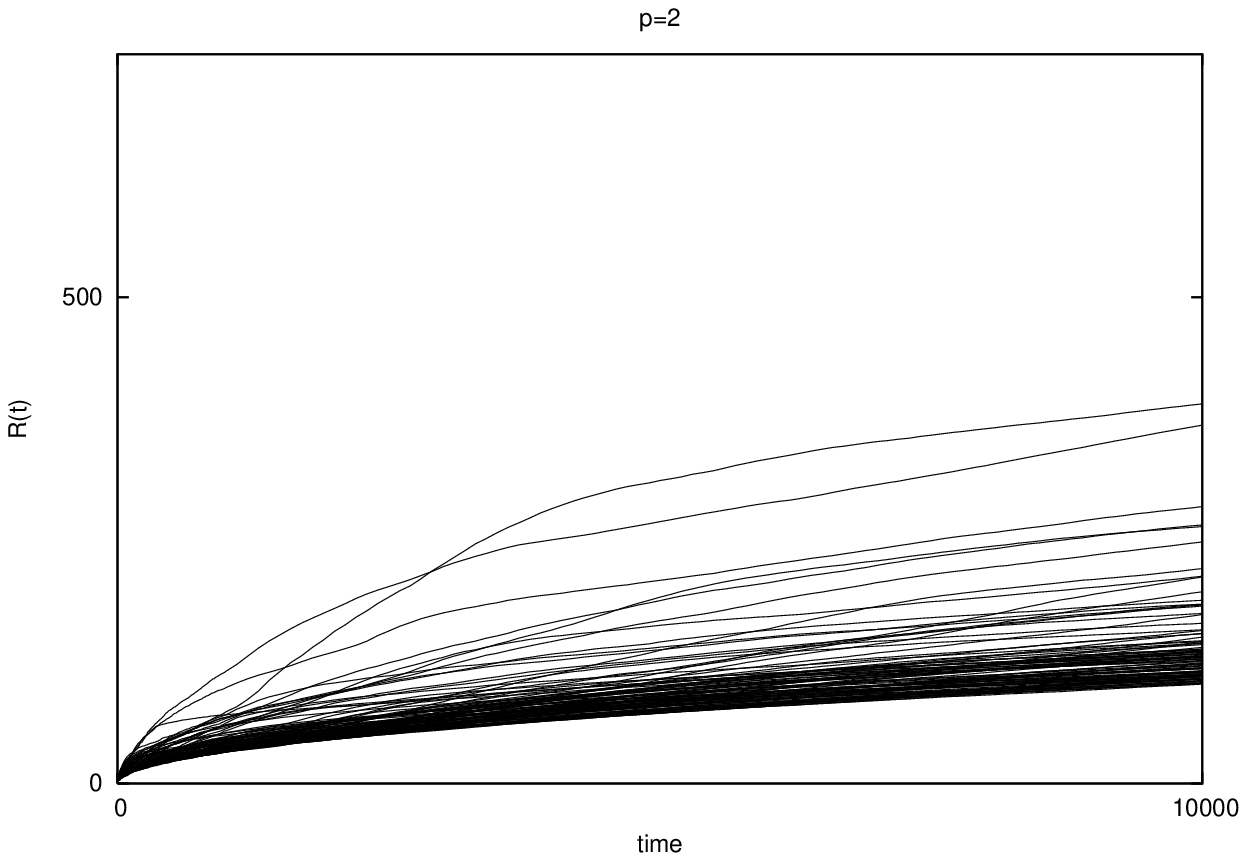}\hfill%
\includegraphics[width=\gwidth]{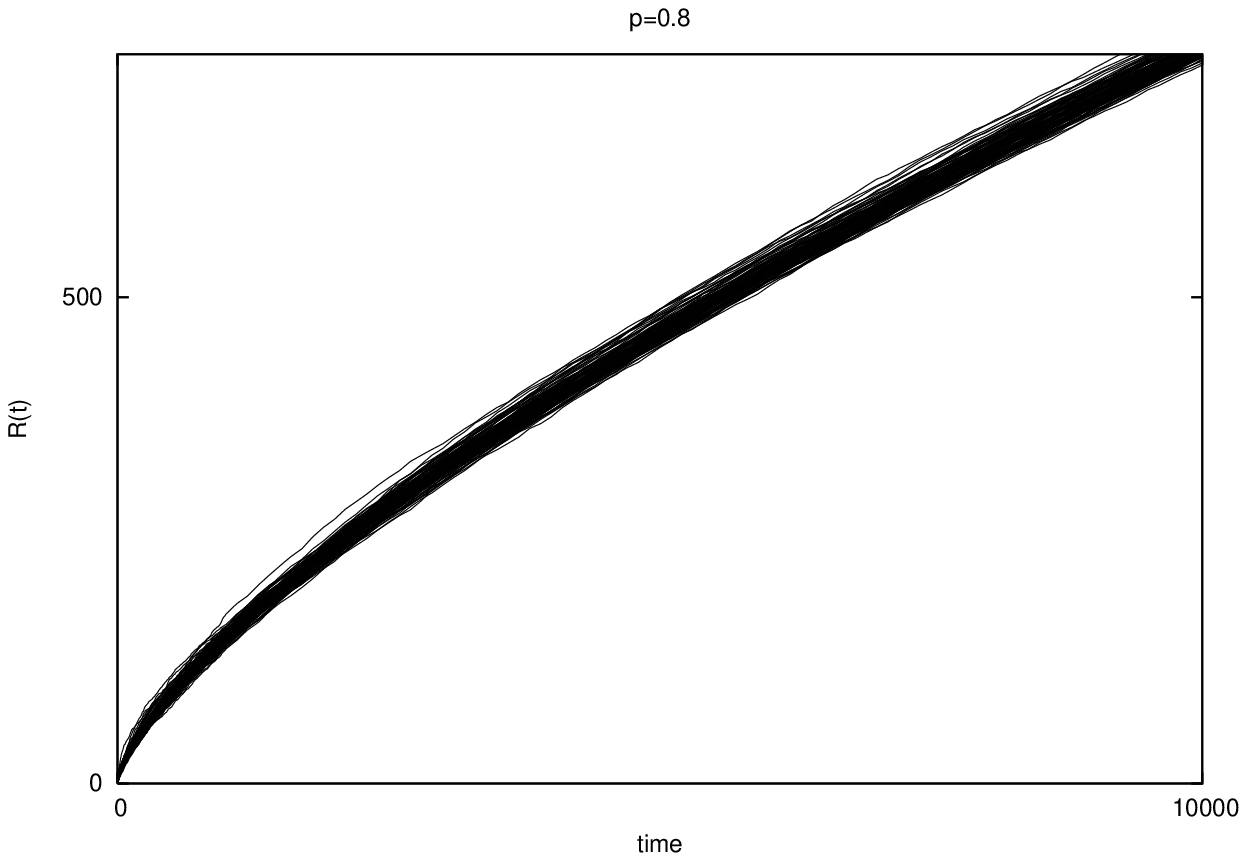}
\caption{Total regret as a function of time for 100 overlaid runs of
  the algorithm of Theorem~\ref{thm:alg} (bottom left) which optimally
  trades off exploration and exploitation; with $p=0.8$ for more query
  noise (bottom right) resulting in less between-run variation but
  more regret; with $p=3.6$ for less query noise (top left) resulting
  in more between-run variation; and for the greedy strategy, zero
  query noise (top right) in which runs rapidly converge to incorrect
  estimates.  All runs used $\sigma^2=a=1$, $b=c=0$, and were
  initialised with two queries at $x = x^* \pm 1$.}
\label{fig:runs}
\end{figure*}

\begin{figure*}[t!]
\centerline{\input{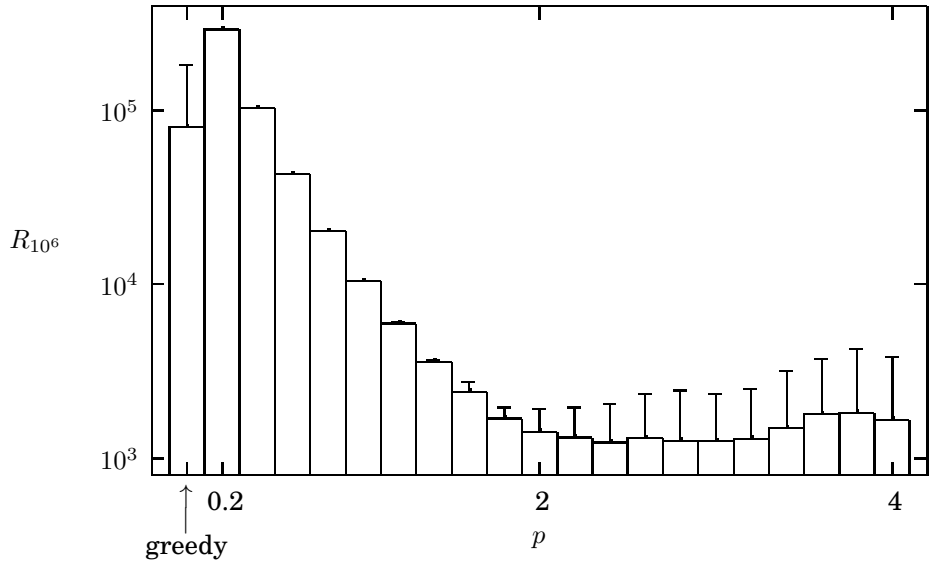}}
\caption{Bar graph (log scale) of total regret after $10^6$ queries,
  averaged over 100 runs, for the algorithm of Theorem~\ref{thm:alg}
  with $\sigma=1$ and $a=1$.  Bars shown for values of $p$ both above
  and below the optimal $p=2$, and also for the greedy algorithm of
  zero injected noise.  Risers show sample standard deviations.}
\label{fig:totals}
\end{figure*}

\section{Discussion}

Although the above theorems all assume unbiased estimates, integration
of prior information would, assuming that the prior is smooth, only
change an initial transient response of the system, leaving the
asymptotic behaviour unchanged.  The limits on regret would change by
only a small additive constant whose value would dependant upon the
details of the prior.

The above exploration/exploitation tradeoff and bound holds when using
noisy measurements and the cost of an evaluation is the value of the
function being optimised.  The result is robust, in that small changes
to the model (a cost function quadratic only in the neighbourhood of
the optimum, for instance) will not change their character.

However a related situation, finding the zero $x^*$ of a linear
function using noisy measurements where the expected loss of a
measurement $x_t$ is quadratic in $x_t - x^*$, has a surprisingly
different result.  In this matching-shoulders lob-pass case formalised
by \citet{ABE-TAKEUCHI93A} based on the foraging theory question posed
by \citet{HERRNSTEIN90A}, a convergence rate of $E[(x_t - x^*)^2] =
O(t^{-1})$ and thus an expected regret of $E[R_t] = O(\log t)$ can be
achieved \citep{KILIAN-ETAL94A, HIRAOKA-AMARI98A,
TAKEUCHI-ETAL-2000a}.  This is because the measurements in that
setting serve the purpose of gradient information.

Procedures which do not insert sufficient variability into their
queries acquire only finite leverage, resulting (with probability one)
in convergence to a non-optimum.  This is seen in the upper
simulations of \fig{runs}.  The minimal total regret in \fig{totals}
is for an algorithm injecting slightly less query than $\stderr
\hat{x}^*_t$.  This is due to the slight additional leverage caused by
fluctuation of the estimate $\hat{x}^*_t$ over time.

Some procedures used in practise for problems of this character appear
to attempt to exceed the convergence bound established here, for
instance in medical treatment optimisation.  The above bounds should
serve as a caution concerning the ease with which a seemingly
reasonable optimisation procedure can converge to a non-optimum.  In
the setting considered here, when insufficient query variance is used
convergence to a non-optimum occurs, and standard statistical analysis
of the ongoing measurements will fail to give any hint of a problem.
Query variability must be injected when the setting itself requires
it, rather than only in response to empirical signs of premature
convergence.

In business, the best selling price (which is not subject to the above
constraint, as noisy \emph{gradient} information is available) should
be faster to estimate than the supply or demand curves, which seem
potentially subject to this bound.  This would argue that firms that
set their prices by first estimating supply and demand curves may be
at a disadvantage against those that set prices directly.  More
speculatively, regulatory regimes have surprising variability
considering that all are designed to further similar goals.  Legal
systems have similar diversity.  The ultimate cause of this
variability may be the intrinsic difficulty of gradient-free noisy
query optimisation.  Even more speculatively, sexual selection for
adaptive traits may provide a proxy for gradient information, thus
speeding evolution.

\subsection*{Acknowledgements}

Supported by Science Foundation Ireland grant 00/PI.1/C067.  Thanks to
Tony Zador, Ken Duffy, and Susanna Still for helpful comments.

\renewcommand{\bibsection}[0]{\subsection*{References}}
\setlength{\bibsep}{1ex}
\setlength{\bibhang}{0.75em}
\bibliographystyle{abbrvnat}    
\bibliography{abb-abbr,boltzmann}

\begin{thebibliography}{8}
\expandafter\ifx\csname natexlab\endcsname\relax\def\natexlab#1{#1}\fi
\expandafter\ifx\csname url\endcsname\relax
  \def\url#1{{\tt #1}}\fi

\bibitem[Abe and Takeuchi(1993)]{ABE-TAKEUCHI93A}
N.~Abe and J.-i. Takeuchi.
\newblock The `lob-pass' problem and an on-line learning model of rational
  choice.
\newblock In {\em Sixth Annual {ACM} Workshop on Computational Learning
  Theory}, pages 422--428, Santa Cruz, CA, July 1993.

\bibitem[Herrnstein(1990)]{HERRNSTEIN90A}
R.~Herrnstein.
\newblock Rational choice theory.
\newblock {\em American Psychologist}, 45\penalty0 (3):\penalty0 356--367,
  1990.

\bibitem[Hiraoka and Amari(1998)]{HIRAOKA-AMARI98A}
K.~Hiraoka and S.-i. Amari.
\newblock Strategy under the unknown stochastic environment: The nonparametric
  lob-pass problem.
\newblock {\em Algorithmica}, 22\penalty0 (1/2):\penalty0 138--156, 1998.

\bibitem[Kilian et~al.(1994)Kilian, Lang, and Pearlmutter]{KILIAN-ETAL94A}
J.~Kilian, K.~J. Lang, and B.~A. Pearlmutter.
\newblock Playing the matching-shoulders lob-pass game with logarithmic regret.
\newblock In {\em Seventh Annual {ACM} Workshop on Computational Learning
  Theory}, pages 159--164, New Brunswick, NJ, July 1994.

\bibitem[Ljung(1977)]{LJUNG77}
L.~Ljung.
\newblock Analysis of recursive stochastic algorithms.
\newblock {\em IEEE Trans. Automat. Contr.}, 22\penalty0 (4):\penalty0
  551--575, 1977.

\bibitem[Robbins and Monro(1951)]{ROBBINS-MONRO51a}
H.~Robbins and S.~Monro.
\newblock A stochastic approximation method.
\newblock {\em Annals of Mathematical Statistics}, 22:\penalty0 400--407, 1951.

\bibitem[Takeuchi et~al.(2000)Takeuchi, Abe, and Amari]{TAKEUCHI-ETAL-2000a}
J.-i. Takeuchi, N.~Abe, and S.-i. Amari.
\newblock The lob-pass problem.
\newblock {\em J. Comput. Syst. Sci.}, 61\penalty0 (3):\penalty0 523--557,
  2000.

\bibitem[Widrow et~al.(1976)Widrow, McCool, Larimore, and
  Johnson]{WIDROW-MCCOOL-LARIMORE-JOHNSON79}
B.~Widrow, J.~M. McCool, M.~G. Larimore, and C.~R. Johnson, Jr.
\newblock Stationary and nonstationary learning characteristics of the {LMS}
  adaptive filter.
\newblock {\em Proceedings of the IEEE}, 64:\penalty0 1151--1162, 1976.

\end{thebibliography}

\end{document}

